\documentclass[runningheads]{llncs}

\usepackage{url}
\usepackage{rotating}
\usepackage{graphicx}
\usepackage{amsmath} 
\usepackage{amsfonts}
\usepackage{algorithmic}
\usepackage[linesnumbered,lined,boxed,commentsnumbered]{algorithm2e}
\usepackage{booktabs}
\usepackage{multirow}
\usepackage{threeparttable}
\usepackage{capt-of}
\usepackage{rotating}
\usepackage{multicol}
\usepackage{makecell}
\usepackage{verbatim}
\usepackage{wrapfig}
\usepackage[dvipsnames,table]{xcolor}
\usepackage[normalem]{ulem}
\usepackage{soul}

\definecolor{darkgreen}{rgb}{0,0.5,0}

\newcommand*{\veci}[1]{\boldsymbol{#1}}

\def\RR{\mathbb{R}}

\def\A{{\cal A}}

\def\D{{\cal D}}

\def\L{{\cal L}}

\def\U{{\cal U}}
\def\V{{\cal V}}

\def\Y{{\cal Y}}
\def\Z{{\cal Z}}

\DeclareMathOperator{\PP}{\mathbb{P}}

\DeclareMathOperator{\dist}{dist}

\SetKwProg{algalg}{fonction}{}{}
\SetKwProg{algalgp}{procédure}{}{}
\SetKwFunction{sgmulti}{SGmulti}{}
\SetKwInput{KwData}{Données}
\SetKwInput{KwResults}{Résultats}
\SetKwInput{KwResult}{Résultat}
\SetKwInput{KwIns}{Entrées}
\SetKwInput{KwOuts}{Sorties}
\SetKwInput{KwIn}{Entrée}
\SetKwInput{KwOut}{Sortie}
\SetKw{KwTo}{jusqu'à}
\SetKw{KwRet}{retourner}
\SetKw{Return}{retourner}
\SetKwBlock{Begin}{début}{fin}
\SetKwComment{tcc}{/*}{*/}
\SetKwComment{tcp}{//}{}
\SetKwIF{If}{ElseIf}{Else}{si}{alors}{sinon si}{sinon}{fin si}
\SetKwSwitch{Switch}{Case}{Other}{switch}{do}{case}{otherwise}{endcase}{endsw}
\SetKwFor{For}{pour}{faire}{fin pour}
\SetKwFor{While}{tant que}{faire}{fin tant que}
\SetKwFor{ForEach}{foreach}{do}{endfch}
\SetKwRepeat{Repeat}{repeat}{until}

\begin{document}

\title{Learning active learning at the crossroads? evaluation and discussion}

\titlerunning{Learning active learning at the crossroads? evaluation and discussion}

\author{L. Desreumaux\inst{1}, V. Lemaire\inst{2}}

\authorrunning{L. Desreumaux and V. Lemaire}

\institute{SAP Labs, Paris, France \and Orange Labs, Lannion, France}

\maketitle

\begin{abstract}
Active learning aims to reduce annotation cost by predicting which samples are useful for a human expert to label. Although this field is quite old, several important challenges to using active learning in real-world settings still remain unsolved. In particular, most selection strategies are hand-designed, and it has become clear that there is no best active learning strategy that consistently outperforms all others in all applications. This has motivated research into meta-learning algorithms for ``learning how to actively learn''. In this paper, we compare this kind of approach with the association of a Random Forest with the margin sampling strategy, reported in recent comparative studies as a very competitive heuristic. To this end, we present the results of a benchmark performed on 20 datasets that compares a strategy learned using a recent meta-learning algorithm with margin sampling. We also present some lessons learned and open future perspectives.
\end{abstract}

\section{Introduction}

Modern supervised learning methods\footnote{In this article, we limit ourselves to binary classification problems.} are known to require large amounts of training examples to reach their full potential. Since these examples are mainly obtained through human experts who manually label samples, the labelling process may have a high cost. Active learning~(AL) is a field that includes all the selection strategies that allow to iteratively build the training set of a model in interaction with a human expert, also called oracle. The aim is to select the most informative examples to minimize the labelling cost.

In this article, we consider the selective sampling framework, in which the strategies manipulate a set of examples $\D = \L \cup \U$ of constant size, where $\L = \{(\veci{x}_i,y_i)\}_{i=1}^l$ is the set of labelled examples and $\U = \{\veci{x}_i\}_{i=l+1}^n$ is the set of unlabelled examples. In this framework, active learning is an iterative process that continues until a labelling budget is exhausted or a pre-defined performance threshold is reached. Each iteration begins with the selection of the most informative example $\veci{x}^{\star}\in\U$. This selection is generally based on information collected during previous iterations (predictions of a classifier, density measures, etc.). The example $\veci{x}^{\star}$ is then submitted to the oracle that returns the corresponding class $y^{\star}$, and the pair $(\veci{x}^{\star}, y^{\star})$ is added to $\L$. The new learning set is then used to improve the model and the new predictions are used in the next iteration.

The utility measures defined by the active learning strategies in the literature~\cite{Settles2012} differ in their positioning according to a dilemma between the exploitation of the current classifier and the exploration of the training data. Selecting an unlabelled example in an unknown region of the observation space $\RR^d$ helps to explore the data, so as to limit the risk of learning a hypothesis too specific to the current set $\L$. Conversely, selecting an example in a sampled region of $\RR^d$ locally refines the predictive model.

\subsection{Traditional heuristic-based AL}\label{firstage}

The active learning field comes from a parallel between active educational methods and machine learning theory. The learner is from now a statistical model and not a student. The interactions between the student and the teacher correspond to the interactions between the model and the oracle. The examples are situations used by the model to generate knowledge on the problem.


The first AL algorithms were designed with the objective of transposing these ``educational'' methods to the machine learning domain. The easiest way was to keep the usual supervised learning methods and to add ``strategies'' relying on various heuristics to guide the selection of the most informative examples. From the first initiative and up to now, a lot of strategies motivated by human intuitions have been suggested in the literature. The purpose of this paper is not to give an overview of the existing strategies but the reader may find in \cite{Settles2012,Aggarwal2014} of lot of them.

A careful reading of the experimental results published in the literature shows that there is no best AL strategy that consistently outperforms all others in all applications, and some strategies cater to specific classifiers or to specific applications. Based on this observation, several comprehensive benchmarks carried out on numerous datasets have highlighted the strategies which, on average, are the most suitable for several classification models \cite{Santos2014,Yang2018,PereiraSantos2019}. They are given in Table~\ref{table:etatart}. For example, the most appropriate strategy for logistic regression and random forest is an uncertainty-based sampling\footnote{The reader interested in the measures used to quantify the degree of uncertainty in the context of active learning may find in \cite{Nguyen2019EpistemicUS,hllermeier2019aleatoric} an interesting view which advocates a distinction between two different types of uncertainty, referred to as epistemic and aleatoric.} strategy, named margin sampling, which consists in selecting at each iteration the instance for which the difference between the probabilities of the two most likely classes is the smallest~\cite{Scheffer2001}. To produce this table, we purposefully omitted studies that have a restricted scope, such as focusing on too few datasets \cite{Beyer2015}, specific tasks \cite{Settles2008}, an insufficient number of strategies \cite{Schein2007,RamirezLoaiza2017}, or variants of a single strategy \cite{Korner2006}.

\begin{table}[htb!]
\centering
\begin{tabular}{cccccccc}
\toprule
\thead{Strategy} & \thead{RF\textsuperscript{1}} & \thead{SVM\textsuperscript{2}} & \thead{5NN\textsuperscript{3}} & \thead{GNB\textsuperscript{4}} & \thead{C4.5\textsuperscript{5}} & \thead{LR\textsuperscript{6}} & \thead{VFDT\textsuperscript{7}} \\ \midrule
Margin\textsuperscript{a} & \cite{PereiraSantos2019} & & & & & & \\
Entropy\textsuperscript{b} & & & & & & \cite{Yang2018} & \\
QBD\textsuperscript{c} & & & & \cite{Santos2014} & & & \cite{Santos2014} \\
Density\textsuperscript{d} & & & \cite{PereiraSantos2019,Santos2014} & & \cite{Santos2014} & & \\
OER\textsuperscript{e} & & \cite{PereiraSantos2019} & & \cite{PereiraSantos2019} & \cite{PereiraSantos2019} & & \\
\bottomrule
\end{tabular}
\caption{Best model/strategy associations highlighted in the literature as a guide to  use the appropriate  strategy versus the classifier. Strategies: (a) Margin sampling, (b) Entropy sampling, (c) Query by Disagreement, (d) Density sampling, (e) Optimistic Error Reduction. Classifiers: (1) Random Forest, (2) Support Vector Machine, (3) 5-Nearest Neighbors, (4) Gaussian Naive Bayes, (5) C4.5 Decision Tree, (6) Logistic Regression, (7) Very Fast Decision Tree.}
\label{table:etatart}
\end{table}

\subsection{Meta-learning approaches to active learning}
\label{secondage}

While the traditional AL strategies can achieve remarkable performance, it is often challenging to predict in advance which strategy is the most suitable in a particular situation. In recent years, meta-learning algorithms have been gaining in popularity \cite{Lemke2015}. Some of them have been proposed to tackle the problem of learning AL strategies instead of relying on manually designed strategies.

Motivated by the success of methods that combine predictors, the first AL algorithms within this paradigm were designed to combine traditional AL strategies with bandit algorithms \cite{Baram2004,Ebert2012,Hsu2015,Chu2016,Collet2018,Pang2018}. These algorithms learn how to select the best AL criterion for any given dataset and adapt it over time as the learner improves. However, all the learning must be achieved within a few examples to be helpful, and these algorithms suffer from a cold start issue. Moreover, these approaches are restricted to combining existing AL heuristic strategies.

Within the meta-learning framework, some other algorithms have been developed to learn from scratch an AL strategy on multiple source datasets and transfer it to new target datasets \cite{Konyushkova2017,Konyushkova2018,Pang2018b}. Most of them are based on modern reinforcement learning methods. The key challenge consists in learning an AL strategy that is general enough to automatically control the exploitation/exploration trade-off when used on new unlabelled datasets, which is not possible when using heuristic strategies.


\subsection{Objective of this paper}
From the state of the art, it appears that meta-learned AL strategies can outperform the most widely used traditional AL strategies, like uncertainty sampling. However, most of the papers that introduce new meta-learning algorithms do not include comprehensive benchmarks that could ascertain the transferability of the learned strategies and demonstrate that these strategies can safely be used in real-world settings.

The objective of this article is thus to compare two possible options in the realization of an AL solution that could be used in an industrial context: using a traditional heuristic-based strategy (see Section \ref{firstage}) that, on average, is the best one for a given model and could be used as a strong baseline easy to implement and not so easy to beat, or using a more sophisticated strategy learned in a data-driven fashion that comes from the very recent literature on meta-learning (see Section \ref{secondage}).

To this end, we present the results of a benchmark performed on 20 datasets that compares a strategy learned using the meta-learning algorithm proposed in~\cite{Konyushkova2018} with margin sampling \cite{Scheffer2001}, the models used being in both cases logistic regression and random forest. We evaluated the work of \cite{Konyushkova2018} since the authors claim to be able to learn a ``general-purpose" AL strategy that can generalise across diverse problems and outperform the best heuristic and bandit approaches.

The rest of the paper is organized as follows. In Section \ref{section:description}, we explain all the aspects of the Learning Active Learning~(LAL) method proposed in \cite{Konyushkova2018}, namely the Deep Q-Learning algorithm and the modeling of active learning as a Markov decision process~(MDP). In Section \ref{section:protocol}, we present the protocol used to do extensive comparative experiments on public datasets from various application areas. In Section \ref{section:results}, we give the results of our experimental study and make some observations. Finally, we present some lessons learned and we open future perspectives in Section \ref{section:conclu}.


\section{Learning active learning strategies} \label{section:description}

\subsection{Q-Learning}

A Markov decision process is a formalism for modeling the interaction between an agent and its environment. This formalism uses the concepts of \textit{state}, which describes the situation in which the environment finds itself, \textit{action}, which describes the decision made by the agent, and \textit{reward}, received by the agent when it performs an action. The procedure followed by the agent to select the action to be performed at time $t$ is the \textit{policy}. Given a policy $\pi$, the \textit{state-action table} is the function $Q^{\pi}(s,a)$ which gives the expectation of the weighted sum of the rewards received from the state $s$ if the agent first executes the action $a$ and then follows the policy $\pi$.

Q-Learning is a reinforcement learning algorithm that estimates the optimal state-action table $Q^{\star} = \max_{\pi} Q^{\pi}$ from interactions between the agent and the environment. The state-action table $Q$ is updated at any time from the current state $s$, the action $a = \pi(s)$ where $\pi$ is the policy derived from $Q$, the reward received $r$ and the next state of the environment $s'$:
\begin{equation}
Q_{t+1}(s, a) = (1-\alpha_t(s,a)) Q_t(s, a) + \alpha_t(s,a) \left(r + \gamma \max_{a'\in\A} Q_t(s', a') \right),
\end{equation}
where $\gamma\in[0,1[$ is the weighting factor of the rewards and the $\alpha_t(s,a)\in\, ]0,1[$ are the learning steps that determine the weight of the new experience in relation to the knowledge acquired at previous steps. Assuming that all the state-action pairs are visited an infinite number of times and under some conditions on the learning steps, the resulting sequence of state-action tables converges to $Q^{\star}$ \cite{Watkins1992}.

The goal of a reinforcement learning agent is to maximize the rewards received over the long term. To do this, in addition to actions that seem to lead to high rewards (exploitation), the agent must select potentially suboptimal actions that allow him to acquire new knowledge about the environment (exploration). For Q-Learning, the $\epsilon$-greedy method is the most commonly used to manage this dilemma. It consists in randomly exploring with a probability of $\epsilon$ and acting according to a greedy strategy that chooses the best action with a probability of $(1-\epsilon)$. It is also possible to decrease the probability $\epsilon$ at each transition to model the fact that exploration becomes less and less useful with time.

\subsection{Deep Q-Learning}\label{section:deepql}

In the Q-Learning algorithm, if the state-action table is implemented as a two-input table, then it is impossible to deal with high-dimensional problems. It is necessary to use a parametric model that will be noted as $Q(s,a;\veci{\theta})$. If it is a deep neural network, it is called Deep Q-Learning.

The training of a neural network requires the prior definition of an error criterion to quantify the loss between the value returned by the network and the actual value. In the context of Q-Learning, the latter value does not exist: one can only use the reward obtained after the completion of an action to calculate a new value, and then estimate the error achieved by calculating the difference between the old value and the new one. A possible cost function would thus be the following:
\begin{equation}
L(s,a,r,s',\veci{\theta}) = \left(r + \gamma \max_{a'\in\A} Q(s', a'; \veci{\theta}) - Q(s, a;\veci{\theta})\right)^2.
\end{equation}
However, this poses an obvious problem: updating the parameters leads to updating the target. In practice, this means that the training procedure does not converge.

In 2013, a successful implementation of Deep Q-Learning introducing several new features was published \cite{Mnih2013}. The first novelty is the introduction of a target network, which is a copy of the first network that is regularly updated. This has the effect of stabilizing learning. The cost function becomes:
\begin{equation}
L(s,a,r,s',\veci{\theta},\veci{\theta}^{-}) = \left(r + \gamma \max_{a'\in\A} Q(s', a'; \veci{\theta}^{-}) - Q(s, a;\veci{\theta})\right)^2,
\end{equation}
where $\veci{\theta}^{-}$ is the vector of the target network parameters. The second novelty is experience replay. It consists in saving each experience of the agent $(s_i, a_i, r_i, s_{i+1})$ in a memory of size $m$ and using random samples drawn from it to update the parameters by stochastic gradient descent. This random draw allows to not necessarily select consecutive, potentially correlated experiences.

\subsection{Improvements to Deep Q-Learning}\label{section:ameliorations}

Many improvements to Deep Q-Learning have been published since the article that introduced it. We present here the improvements that interest us for the study of the LAL method.

\paragraph{Double Deep Q-Learning.}
A first improvement is the correction of the overestimation bias. It has indeed been empirically shown that Deep Q-Learning as presented in Section~\ref{section:deepql} can produce a positive bias that increases the convergence time and has a significant negative impact on the quality of the asymptotically obtained policy. The importance of this bias and its consequences have been verified in particular in the configurations the least favourable to its emergence, \textit{i.e.} when the environment and rewards are deterministic. In addition, its value increases with the size of the set of actions. To correct this bias, the solution which has been proposed in \cite{Hasselt2016} consists in not using the parameters $\veci{\theta}^{-}$ to both select and evaluate an action. The cost function then becomes:
\begin{equation}
L(s,a,r,s',\veci{\theta},\veci{\theta}^{-}) = \left(r + \gamma Q\left(s', \arg\max_{a'\in\A} Q(s',a';\veci{\theta});\veci{\theta}^{-}\right) - Q(s, a;\veci{\theta})\right)^2.
\end{equation}

\paragraph{Prioritized Experience Replay.}
Another improvement is the introduction of the notion of priority in experience replay. In its initial version, Deep Q-Learning considers that all the experiences can identically advance learning. However, reusing some experiences at the expense of others can reduce the learning time. This requires the ability to measure the acceleration potential of learning associated with an experience. The priority measure proposed in \cite{Schaul2016} is the absolute value of the temporal difference error:
\begin{equation}
\delta_i = \left| r_i + \gamma \max_{a'\in\A} Q(s_{i+1},a'; \veci{\theta}^{-}) - Q(s_i,a_i;\veci{\theta}) \right|.
\end{equation}
A maximum priority is assigned to each new experience, so that all the experiences are used at least once to update the parameters. 

However, the experiences that produce a small temporal difference error at first use may never be reused. To address this issue, a method was introduced in~\cite{Schaul2016} to manage the trade-off between uniform sampling and sampling focusing on experiences producing a large error. It consists in defining the probability of selecting an experience $i$ as follows: 
\begin{equation}\label{eq:beta}
p_i = \frac{\rho_i^{\beta}}{\sum_{k=1}^{m} \rho_k^{\beta}},\quad\text{with}\quad \rho_i = \delta_i + e,
\end{equation}
where $\beta\in\RR^+$ is a parameter that determines the shape of the distribution and $e$ is a small positive constant that guarantees $p_i > 0$. The case where $\beta = 0$ is equivalent to uniform sampling.

\subsection{Formulating active learning as a Markov decision process}\label{section:actifmdp}

The formulation of active learning as a MDP is quite natural. In each MDP \textit{state}, the \textit{agent} performs an \textit{action}, which is the selection of an instance to be labelled, and the latter receives a \textit{reward} that depends on the quality of the model learned with the new instance. The active learning strategy becomes the MDP \textit{policy} that associates an action with a state.

In this framework, the iteration $t$ of the policy learning process from a dataset divided into a learning set $\D = \L_t \cup \U_t$ and a test set\footnote{Given that active learning is usually applied in cases, this test set assumed to be small or very small the performance evaluated on this test set could be a possibly bad approximation. This issue and techniques for avoiding it are not examined in this paper.} $\D'$ consists in the following steps:
\begin{enumerate}
    \item A model $h^{(t)}$ is learned from $\L_t$. Associated with $\L_t$ and $\U_t$, it allows to characterize a state $\veci{s}_t$.
    \item The agent performs the action $\veci{a}_t = \pi(\veci{s}_t)\in \A_t$ which defines the instance $\veci{x}^{(t)}{\in \U_t}$ to label.
    \item The label $y^{(t)}$ associated with $\veci{x}^{(t)}$ is retrieved and the training set is updated, \textit{i.e. } $\L_{t+1} = \L_t \cup \{ (\veci{x}^{(t)}, y^{(t)})\}$ and $\U_{t+1} = \U_t \setminus \{\veci{x}^{(t)} \}$.
    \item The agent receives the reward $r_{t}$ associated with the performance $\ell_t$ on the test set $\D'$. This reward is used to update the policy (see Section~\ref{section:apprentissagedqn}).
\end{enumerate}
The set of actions $\A_t$ depends on time because it is not possible to select the same instance several times. These steps are repeated until a terminal state $s_T$ is reached. Here, we consider that we are in a terminal state when all the instances have been labelled or when $\ell_t \geq q$, where $q$ is a performance threshold that has been chosen as 98\% of the performance obtained when the model is learned on all the training data.

The precise definition of the set of states, the set of actions and the reward function is not evident. To define a state, it has been proposed to use a vector whose components are the scores $\widehat{y}_t(\veci{x}) = \PP(\mathrm{Y} = 0\, |\, \veci{x})$ associated with the unlabelled instances of a subset $\V$ set aside. This is the simplest representation that can be used to characterize the uncertainty of a classifier on a dataset at a given time $t$.

The set of actions has been defined at iteration $t$ as the set of vectors $\veci{a}_i =[\widehat{y}_t(\veci{x}_i), g(\veci{x}_i, \L_t), g(\veci{x}_i, \U_t)]$, where $\veci{x}_i\in\U_t$ and :
\begin{equation}
g(\veci{x}_i, \L_t) = \frac{1}{|\L_t|}\sum_{\veci{x}_j\in\L_t} \dist(\veci{x}_i, \veci{x}_j),\quad g(\veci{x}_i, \U_t) = \frac{1}{|\U_t|}\sum_{\veci{x}_j\in\U_t} \dist(\veci{x}_i, \veci{x}_j),
\end{equation}
where $\dist$ is the cosine distance. An action is therefore characterized by the uncertainty on the associated instance, as well as by two statistics related to the density of the neighbourhood of the instance.

The reward function has been chosen constant and negative until arrival in a terminal state ($r_t = -1$). Thus, to maximize its reward, the agent must perform as few interactions as possible.

\subsection{Learning the optimal policy through Deep Q-Learning}\label{section:apprentissagedqn}

The Deep Q-Learning algorithm with the improvements presented in Section~\ref{section:ameliorations} is used to learn the optimal policy. To be able to process a state space that evolves with each iteration, the neural network architecture has been modified. The new architecture considers actions as inputs to the $Q$ function in the same way as states. It then returns only one value, while the classical architecture takes only one state as input and returns the values associated with all the actions.

\begin{figure}[htbp]
    \centering
    \includegraphics[width=\linewidth]{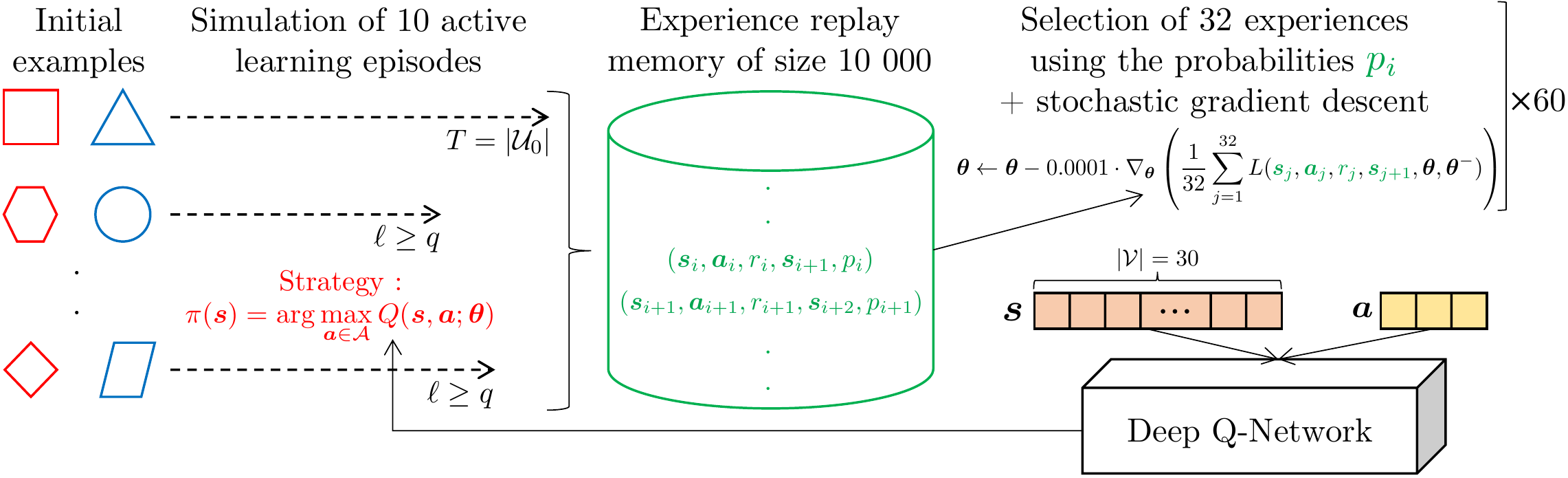}
    \caption{Illustration of the different steps involved in an iteration of the policy learning phase using Deep Q-Learning (the arrows give intuitions about main steps and data flows)}.
    \label{figure:politique}
\end{figure}

The learning procedure involves a collection of $Z$ labelled datasets $\{\Z_i \}_{1\leq i\leq Z}$. It consists in repeating the following steps (see Figure~\ref{figure:politique}):
\begin{enumerate}
    \item A dataset $\Z\in\{\Z_i\}$ is randomly selected and divided into a training set $\D$ and a test set $\D'$.
    \item The policy $\pi$ derived from the Deep Q-Network is used to simulate several active learning episodes on $\Z$ according to the procedure described in Section~\ref{section:actifmdp}. Experiences $(\veci{s}_t, \veci{a}_t, r_t, \veci{s}_{t+1})$ are collected in a finite size memory.
    \item The Deep Q-Network parameters are updated several times from a minibatch of experiences extracted from the memory (according to the method described in Section~\ref{section:ameliorations}).
\end{enumerate}

To initialize the Deep Q-Network, some warm start episodes are simulated using a random sampling policy, followed by several parameter updates. Once the strategy is learned, its deployment is very simple. At each iteration of the sampling process, the classifier is re-trained, then the vector characterizing the process state and all the vectors associated with the actions are calculated. The vector $\veci{a}^{\star}$ corresponding to the example to label $\veci{x}^{\star}$ is then the one that satisfies $\veci{a}^{\star} = \arg\max_{\veci{a}\in\A} Q(\veci{s},\veci{a};\veci{\theta})$, the parameters $\veci{\theta}$ being set at the end of the policy learning procedure.

\section{Experimental protocol}\label{section:protocol}

In this section, we introduce our protocol of the comparative experimental study we conducted.

\subsection{Policy learning}

To learn the strategy, we used the same code\footnote{\url{https://github.com/ksenia-konyushkova/LAL-RL}}, the same hyperparameters and the same datasets as those used in \cite{Konyushkova2018}. The complete list of hyperparameters is given in Table~\ref{table:hyperparameters} with the variable names from the code that represent them. The datasets from which the strategy is learned are given in Table~\ref{table:datasetsApprentissageLAL}.

The specification of the neural network architecture is very simple (all the layers are fully connected): (i) the first layer (linear + sigmoid) receives the vector $\veci{s}$ (\textit{i.e.} $|\V|=30$ input neurons) and has 10 output neurons; (ii) the second layer (linear + sigmoid) concatenates the 10 output neurons of the first layer with the vector $\veci{a}$ (\textit{i.e.} 13 neurons in total) and has 5 output neurons; (iii) finally, the last layer (linear) has only one output to estimate $Q(\veci{s},\veci{a})$.

\begin{table}[htb!]
\centering
\resizebox{\columnwidth}{!}{
\begin{threeparttable}
\begin{tabular}{ll}
\toprule
\thead{Hyperparameter} & \thead{Description} \\ \midrule
\texttt{N\_STATE\_ESTIMATION = 30} & Size of $\V$ \\
\texttt{REPLAY\_BUFFER\_SIZE = 10000} & Experience replay memory size \\
\texttt{PRIORITIZED\_REPLAY\_EXPONENT = 3} & Exponent $\beta$ involved in Equation~\eqref{eq:beta} \\
\texttt{BATCH\_SIZE = 32} & Minibatch size for stochastic gradient descent \\
\texttt{LEARNING\_RATE = 0.0001} & Learning rate \\
\texttt{TARGET\_COPY\_FACTOR = 0.01} & Value that sets the target network update\tnote{1} \\
\texttt{EPSILON\_START = 1} & Exploration probability at start \\
\texttt{EPSILON\_END = 0.1} & Minimum exploration probability \\
\texttt{EPSILON\_STEPS = 1000} & Number of updates of $\epsilon$ during the training \\
\texttt{WARM\_START\_EPISODES = 100} & Number of warm start episodes \\
\texttt{NN\_UPDATES\_PER\_WARM\_START = 100} & Number of parameter updates after the warm start \\
\texttt{TRAINING\_ITERATIONS = 1000} & Number of training iterations \\
\texttt{TRAINING\_EPISODES\_PER\_ITERATION = 10} & Number of episodes per training iteration \\
\texttt{NN\_UPDATES\_PER\_ITERATION = 60} & Number of updates per training iteration \\
\bottomrule
\end{tabular}
\begin{tablenotes}
\item[1] In this implementation, the target network parameters $\veci{\theta}^{-}$ are updated each time the parameters $\veci{\theta}$ are changed as follows: $ \veci{\theta}^{-} \leftarrow (1 - \texttt{TARGET\_COPY\_FACTOR}) \cdot \veci{\theta}^{-} + \texttt{TARGET\_COPY\_FACTOR} \cdot \veci{\theta}.$ \end{tablenotes}
\end{threeparttable}
}
\vspace{0.01cm}
\caption{Hyperparameters involved in Deep Q-Learning.}
\label{table:hyperparameters}
\vspace{-4mm}
\end{table}

\begin{table}[htb!]
\centering
\fontsize{8}{10}\selectfont
\begin{tabular}{lrrrrrr}
\toprule
\thead{Dataset} & \thead{$|\D|$} & \thead{$|\Y|$} & \thead{\#num} & \thead{\#cat} & \thead{maj (\%)} & \thead{min (\%)} \\ \midrule
australian & 690 & 2 & 6 & 8 & 55.51 & 44.49 \\
breast-cancer & 272 & 2 & 0 & 9 & 70.22	& 29.78 \\
diabetes & 768 & 2 & 8 & 0 & 65.10 & 34.90 \\
german & 1000 & 2 & 7 & 13 & 70.00 & 30.00 \\
heart & 293 & 2 & 13 & 0 & 63.82 & 36.18 \\
ionosphere & 350 & 2 & 33 & 0 & 64.29 & 35.71 \\
mushroom & 8124 & 2 & 0 & 21 & 51.80 & 48.20 \\
wdbc & 569 & 2 & 30 & 0 & 62.74 & 37.26 \\ \bottomrule
\end{tabular}
\caption{Datasets used to learn the new strategy. Columns: number of examples, number of classes, numbers of numerical and categorical variables, proportions of examples in the majority and minority classes.}
\label{table:datasetsApprentissageLAL}
\end{table}

\subsection{Traditional heuristic-based AL used as baseline: margin sampling}
Our objective is to compare the performance of a strategy learned using LAL with the performance of a heuristic strategy that, on average, is the best one for a given model. Several benchmarks conducted on numerous datasets have highlighted the fact that margin sampling is the best heuristic strategy for logistic regression (LR) and random forest (RF)~\cite{Yang2018,PereiraSantos2019}. 

Margin sampling consists in choosing the instance for which the difference (or margin) between the probabilities of the two most likely classes is the smallest:
\begin{equation}
\veci{x}^{\star} = \arg\min_{\veci{x}\in\U} \PP(y_1\, |\, \veci{x}) - \PP(y_2\, |\, \veci{x}),
\end{equation}
where $y_1$ and $y_2$ are respectively the first and second most probable classes for $\veci{x}$. The main advantage of this strategy is that it is easy to implement: at each iteration, a single training of the model and $|\U|$ predictions are sufficient to select an example to label. A major disadvantage, however, is its total lack of exploration, as it only exploits locally the hypothesis learned by the model.

We chose to evaluate the Margin/LR association because it is with logistic regression that the hyperparameters of Table~\ref{table:hyperparameters} were optimized in \cite{Konyushkova2018}. In addition, in order to determine whether it is necessary to modify them when another model is used, we also evaluated the Margin/RF association. This last association is particularly interesting because it is the best association highlighted in a recent and large benchmark carried out on 73 datasets, including 5 classification models and 8 active learning strategies~\cite{PereiraSantos2019}. In addition, we evaluated random sampling (Rnd) for both models.

\subsection{Datasets}

The datasets were selected so as to have a high diversity according to the following criteria: (i) number of examples; (ii) number of numerical variables; (iii) number of categorical variables; (iv) class imbalance.

We have also taken care to exclude datasets that are too small and not representative of those used in an industrial context. The 20 selected datasets are described in Table~\ref{table:datasetsEvaluationLAL}. They all come from the UCI database \cite{Dua:2019}, apart from the dataset ``orange-fraud'' which is dataset on fraud detection. Four of the datasets have been used in a challenge on active learning that took place in 2010~\cite{Guyon2011} and the dataset
``nomao'' comes from another challenge on active learning~\cite{Candillier2013}.

\begin{table}[htbp]
\centering
\fontsize{8}{10}\selectfont
\begin{tabular}{lrrrrrr}
\toprule
\thead{Dataset} & \thead{$|\D|$} & \thead{$|\Y|$} & \thead{\#num} & \thead{\#cat} & \thead{maj (\%)} & \thead{min (\%)} \\ \midrule
adult & 48790 & 2 & 6 & 8 & 76.06 & 23.94 \\
banana & 5292 & 2 & 2 & 0 & 55.16 & 44.84 \\
bank-marketing-full & 45211 & 2 & 7 & 9 & 88.30 & 11.70 \\
climate-simulation-craches & 540 & 2 & 20 & 0 & 91.48 & 8.52 \\
eeg-eye-state & 14980 & 2 & 14 & 0 & 55.12 & 44.88 \\
hiva & 40764 & 2 & 1617 & 0 & 96.50 & 3.50 \\
ibn-sina & 13951 & 2 & 92 & 0 & 76.18 & 23.82 \\
magic & 18905 & 2 & 10 & 0 & 65.23 & 34.77 \\
musk & 6581 & 2 & 166 & 1 & 84.55 & 15.45 \\
nomao & 32062 & 2 & 89 & 29 & 69.40 & 30.60 \\
orange-fraud & 1680 & 2 & 16 & 0 & 63.75 & 36.25 \\
ozone-onehr & 2528 & 2 & 72 & 0 & 97.11 & 2.89 \\
qsar-biodegradation & 1052 & 2 & 41 & 0 & 66.35 & 33.65 \\
seismic-bumps & 2578 & 2 & 14 & 4 & 93.41 & 6.59 \\
skin-segmentation & 51444 & 2 & 3 & 0 & 71.51 & 28.49 \\
statlog-german-credit & 1000 & 2 & 7 & 13 & 70.00 & 30.00 \\
thoracic-surgery & 470 & 2 & 3 & 13 & 85.11 & 14.89 \\
thyroid-hypothyroid & 3086 & 2 & 7 & 18 & 95.43 & 4.57 \\
wilt & 4819 & 2 & 5 & 0 & 94.67 & 5.33 \\
zebra & 61488 & 2 & 154 & 0 & 95.42 & 4.58 \\ \bottomrule
\end{tabular}
\caption{Datasets used for the evaluation of the strategy learned by LAL. Columns: number of examples, number of classes, numbers of numerical and categorical variables, proportions of examples in the majority and minority classes.}
\label{table:datasetsEvaluationLAL}
\end{table}

\subsection{Evaluation criteria}

In our evaluation protocol, the active sampling process begins with the random selection of one instance in each class and ends when 250 instances are labelled. This value ensures that our results are comparable to other studies in the literature. For performance comparison, we used the area under the learning curve (ALC) based on the classification accuracy. We do not claim that the ALC is a ``perfect metric"\footnote{There is literature on more expressive summary statistics of the active-learning curve \cite{Trittenbach2018,Pupo2018StatisticalCO}. This could be a limitation of this current article, other metrics could be tested in future versions of  experiments.} but it is the defacto standard evaluation criterion in active learning, and it has been chosen as part of a challenge~\cite{Guyon2011}.

Our evaluation was carried out by cross-validation with 5 partitions, in which class imbalance within the complete dataset was preserved. For each partition, the sampling process was repeated 5 times with different initializations to get a mean and a variance on the result. However, we have made sure that the initial instances are identical for all the strategy/model associations on each partition so as to not introduce bias into the results. In addition, for Rnd, the random sequence of numbers was identical for all the models.

\section{Results}\label{section:results}

The results of our experimental study are given in Table~\ref{table:lalres}. The mean ALC obtained for each dataset/classifier/strategy association are reported (the optimal score is 100). The left part of the table gives the results for logistic regression and the right part gives the results for random forest. The penultimate line corresponds to the averages calculated on all the datasets and the last line gives the number of times the strategy has won, tied or lost. The non-significant differences were established on the basis of a paired $t$-test at 99\% significance level (where H0: same mean between populations and  where the mean is the estimate out of 5 repetitions x  cross-validation with 5 partitions of each method).

\begin{table}[htbp]
\centering
\fontsize{7}{9}\selectfont
\begin{tabular}{l|ccc|ccc|c}
\toprule
\thead{Dataset} & \thead{Rnd/LR} & \thead{Margin/LR} & \thead{LAL/LR} & \thead{Rnd/RF} & \thead{Margin/RF} & \thead{LAL/RF} & \thead{maj} \\ \midrule
adult & 77.93 & 78.91 & 78.97 & 80.17 & {\textbf{81.27}} & {\textbf{81.21}} & 76.06\\
banana & 53.03 & {\textbf{57.39}} & 53.12 & {\textbf{80.24}} & 73.81 & 73.58 & 55.16\\
bank-marketing-full & 86.85 & {\textbf{87.62}} & {\textbf{87.72}} & 88.19 & 88.34 & 88.49 & 88.30\\
climate-simulation & 87.22 & {\textbf{89.13}} & {\textbf{88.62}} & 91.15 & 91.14 & 91.13 & 91.48\\ 
eeg-eye-state & 56.08 & 55.32 & 56.11 & 65.53 & {\textbf{67.58}} & 64.42 & 55.12\\
hiva & 64.43 & {\textbf{70.84}} & {\textbf{71.80}} & 96.32 & {\textbf{96.47}} & {\textbf{96.44}} & 96.50\\
ibn-sina & 84.77 & {\textbf{88.58}} & {\textbf{88.90}} & 90.53 & {\textbf{93.41}} & {\textbf{92.75}} & 76.18\\
magic & 76.49 & {\textbf{77.93}} & {\textbf{77.64}} & 78.05 & {\textbf{80.79}} & {\textbf{79.68}} & 65.23 \\
musk & 83.73 & 82.34 & 81.95 & 89.55 & {\textbf{96.18}} & 95.35 & 84.55\\
nomao & 89.45 & {\textbf{91.43}} & {\textbf{91.37}} & 89.41 & {\textbf{92.32}} & {\textbf{92.07}} & 69.40\\
orange-fraud & 76.70 & {\textbf{81.74}} & 74.26 & 89.15 & {\textbf{90.66}} & {\textbf{90.48}} & 63.75\\
ozone-onehr & 92.90 & 94.26 & {\textbf{95.06}} & 96.61 & 96.83 & 96.89 & 97.11\\
qsar-biodegradation & 80.98 & {\textbf{82.62}} & {\textbf{83.53}} & 80.34 & {\textbf{82.76}} & {\textbf{82.40}} & 66.35\\
seismic-bumps & 90.87 & {\textbf{92.59}} & {\textbf{92.14}} & 92.48 & {\textbf{92.92}} & {\textbf{93.02}} & 93.41\\
skin-segmentation & 77.05 & {\textbf{82.69}} & {\textbf{83.21}} & 91.51 & {\textbf{95.70}} & {\textbf{95.77}} & 71.51\\
statlog-german-credit & 70.76 & {\textbf{72.12}} & {\textbf{72.34}} & 72.25 & 72.93 & 72.78 & 70.00\\
thoracic-surgery & 83.76 & 83.93 & 82.72 & 83.51 & {\textbf{84.41}} & {\textbf{84.18}} & 85.11\\
thyroid-hypothyroid & 97.21 & {\textbf{97.99}} & {\textbf{97.97}} & 97.75 & {\textbf{98.77}} & {\textbf{98.71}} & 95.43\\
wilt & 93.53 & {\textbf{95.18}} & 92.87 & 94.86 & {\textbf{97.23}} & {\textbf{97.02}} & 94.67\\
zebra & 86.40 & 90.31 & {\textbf{91.36}} & 94.71 & {\textbf{95.54}} & 95.25 & 95.42\\\midrule
Mean & 80.51 & {\textbf{82.65}} & 82.08 & 87.12 & {\textbf{88.45}} & 88.08 & 79.53 \\
win/tie/loss & 0/5/15  & {\textbf{3/15/2}} & 2/15/3 & 1/4/15 & {\textbf{3/16/1}} & 0/16/4 & \\
\bottomrule
\end{tabular}
\caption{Results of the experimental study.}
\label{table:lalres}
\end{table}

Several observations can be made. First of all, it should be noted that the choice of model is decisive: the results of random forest are all better than those of logistic regression. The random forest model learns indeed very well from few data, as highlighted in~\cite{Salperwyck2011}. We can notice that even with random sampling, RF is almost always better than LR, regardless of the strategy used. In addition, using margin sampling with this model allows a significant performance improvement. This model is very competitive in itself because by its nature, it includes terms of exploration and exploitation (see Section \ref{section:conclu} Conclusion about this point).

In addition, the results of the learned strategy clearly show that a good active learning strategy has been learned, since it performs better than random sampling over a large number of datasets. However, the learned strategy is no better than margin sampling. These results are nevertheless very interesting since only 8 datasets were used in the learning procedure.

Finally, the results show a well-known fact about active learning: on very unbalanced datasets, it is difficult to achieve a better performance than random sampling, as shown in the last column of Table~\ref{table:lalres} in which the results obtained by always predicting the majority class are given. The ``cold start'' problem that occurs in active learning, \textit{i.e.} the inability of making reliable predictions in early iterations (when training data is not sufficient), is indeed further aggravated when a dataset has highly imbalanced classes, since the selected samples are likely to belong to the majority class \cite{Shao2019}. However, if the imbalance is known, it may be interesting to associate strategies with a model or criterion appropriate to this case, as illustrated in \cite{Ertekin2007}.

To investigate the ``learning speed'', we show results for different sizes of $\L$ in Table~\ref{table:lalres2}. They lead to similar conclusions and our results for $|\L|=32$ confirm the results of~\cite{Salperwyck2011}. The reader may find all our experimental results on Github\footnote{\url{https://github.com/ldesreumaux/lal_evaluation}}.

\begin{table}[htbp]
\centering
\fontsize{7}{9}\selectfont
\begin{tabular}{l|ccc|ccc|ccc|ccc}
\toprule
& \multicolumn{3}{c|}{$|\L|=32$} & \multicolumn{3}{c|}{$|\L|=64$} & \multicolumn{3}{c|}{$|\L|=128$}  & \multicolumn{3}{c}{$|\L|=250$}\\
\thead{Dataset} & \thead{Rnd} & \thead{Margin} & \thead{LAL} & \thead{Rnd} & \thead{Margin} & \thead{LAL} & \thead{Rnd} & \thead{Margin} & \thead{LAL} & \thead{Rnd} & \thead{Margin} & \thead{LAL}\\ \midrule
adult & 77.95 & 77.88 & 78.16 & 79.72 & 80.51 & 81.05 & 81.13 & 82.79 & 82.48 & 82.12 & 83.55 & 83.40\\
banana & 71.13 & 65.48 & 65.16 & 77.93 & 71.42 & 70.96 & 83.64 & 75.58 & 75.70 & 86.55 & 79.71 & 81.35\\
bank... & 88.05 & 87.90 & 88.10 & 88.29 & 88.38 & 88.54 & 88.43 & 88.82 & 88.90 & 88.75 & 89.21 & 89.35\\
climate... & 91.26 & 91.26 & 91.18 & 91.40 & 91.29 & 91.40 & 91.26 & 91.33 & 91.33 & 91.44 & 91.22 & 91.29\\ 
eeg... & 58.28 & 58.94 & 57.34 & 62.07 & 63.17 & 60.79 & 66.77 & 69.38 & 65.35 & 72.55 & 75.08 & 72.46\\
hiva & 96.36 & 96.52 & 96.49 & 96.36 & 96.55 & 96.54 & 96.46 & 96.57 & 96.56 & 96.49 & 96.65 & 96.65\\
ibn-sina & 86.88 & 91.17 & 89.78 & 90.48 & 93.99 & 92.96 & 92.73 & 94.76 & 94.25 & 93.86 & 95.85 & 95.48\\
magic & 71.99 & 75.63 & 72.95 & 76.85 & 80.20 & 77.26 & 80.15 & 82.71 & 82.01 & 82.42 & 84.53 & 84.43\\
musk & 85.29 & 89.50 & 90.09 & 87.43 & 94.44 & 94.18 & 90.58 & 98.78 & 97.63 & 93.64 & 99.98 & 99.31\\
nomao & 85.92 & 89.35 & 89.37 & 88.92 & 92.46 & 92.09 & 90.85 & 93.69 & 93.33 & 92.36 & 94.52 & 94.37\\
orange... & 88.06 & 90.36 & 90.09 & 89.16 & 90.98 & 90.67 & 90.08 & 91.72 & 91.33 & 90.41 & 91.85 & 91.74\\
ozone... & 96.36 & 96.97 & 97.01 & 96.74 & 97.04 & 97.10 & 96.93 & 97.08 & 97.11 & 97.02 & 97.03 & 97.05\\
qsar... & 75.75 & 78.08 & 76.61 & 79.75 & 82.09 & 81.42 & 81.94 & 84.65 & 84.88 & 84.03 & 86.12 & 86.08\\
seismic... & 92.39 & 93.21 & 93.19 & 92.42 & 93.28 & 93.19 & 92.52 & 93.26 & 93.20 & 93.14 & 93.08 & 93.28\\
skin... & 86.42 & 89.19 & 89.46 & 90.80 & 96.19 & 96.06 & 93.70 & 98.86 & 98.65 & 95.85 & 99.56 & 99.49\\
statlog... & 70.36 & 70.70 & 69.70 & 70.94 & 72.47 & 71.75 & 72.40 & 73.46 & 74.10 & 74.29 & 75.22 & 75.06\\
thoracic... & 83.14 & 84.42 & 84.12 & 83.31 & 85.02 & 84.76 & 83.70 & 84.89 & 84.68 & 84.21 & 84.51 & 84.68\\
thyroid... & 97.26 & 98.71 & 98.43 & 97.86 & 99.15 & 98.71 & 98.08 & 99.10 & 98.89 & 98.26 & 98.84 & 98.98\\
wilt & 94.60 & 96.23 & 95.98 & 95.01 & 97.47 & 96.90 & 95.30 & 98.21 & 97.64 & 96.07 & 98.51 & 98.37\\
zebra & 94.66 & 95.32 & 95.28 & 94.87 & 95.44 & 95.31 & 94.96 & 95.72 & 95.46 & 95.01 & 96.04 & 95.33 \\ \midrule
Mean & 84.60 & \textbf{85.84} & 85.42 & 86.51 & \textbf{88.07} & 87.58 & 88.08 & \textbf{89.56} & 89.17 & 89.42 & \textbf{90.55} & 90.40 \\
\bottomrule
\end{tabular}
\caption{Mean test accuracy (\%) for different sizes of $|\L|$ with the random forest model.}
\label{table:lalres2}
\vspace{-4mm}
\end{table}

\section{Discussion and open questions}
\label{section:conclu}
In this article, we evaluated a method representative of a recent orientation of active learning research towards meta-learning methods for ``learning how to actively learn'', which is on top of the state of the art~\cite{Konyushkova2018}, versus a traditional heuristic-based Active Learning (the association of Random Forest and Margin) which is one of the best method reported in recent comparative studies~\cite{Yang2018,PereiraSantos2019}. The comparison is limited to just one representative of each of the two classes (meta-learning and traditional heuristic-based) but since each is on top of the state of the art several lessons can be drawn from our study.

\paragraph{Relevance of LAL.}
First of all, the experiments carried out confirm the relevance of the LAL method, since it has enabled us to learn a strategy that achieves the performance of a very good heuristic, namely margin sampling, but contrary to the results in~\cite{Konyushkova2018}, the strategy is not always better than random sampling. This method still raises many problems, including that of the transferability of the learned strategies. An active learning solution that can be used in an industrial context must perform well on real data of an unknown nature and must not involve parameters to be adjusted. With regard to the LAL method, a first major problem is therefore the constitution of a ``dataset of datasets'' large and varied enough to learn a strategy that is effective in very different contexts. 

Moreover, the learning procedure is sensitive to the performance criteria used, which in our view seems to be a problem. Ideally, the strategy learned should be usable on new datasets with arbitrary performance criteria (AUC, F-score, etc.). From our point of view, the work of optimizing the many hyperparameters of the method (see Table~\ref{table:hyperparameters}) can not be carried out by a user with no expertise in deep reinforcement learning.

\paragraph{About the Margin/RF association.}
In addition to the evaluation of the LAL method, we confirmed a result of~\cite{PereiraSantos2019}, namely that margin sampling, associated with a random forest, is a very competitive strategy. From an industrial point of view, regarding the computational complexity, the performances obtained and the absence of  ``domain knowledge required to be used" the Margin/RF association remains a very strong baseline difficult to beat.
However, it shares a major drawback with many active learning strategies, that is its lack of reliability. Indeed, there is no strategy that is better or equivalent to random sampling on {\bf all} datasets and with all models. The literature on active learning is incomplete with regard to this problem, which is nevertheless a major obstacle to using active learning in real-world settings. 

Another important problem in real-world applications, little studied in the literature, is the estimation of the generalization error without a test set. It would be interesting to check if the Out-Of-Bag samples of the random forests \cite{oob} can be used in an active learning context to estimate this error.

Concerning the exploitation/exploration dilemma, margin sampling clearly performs only exploitation. The good results of the Margin/RF association may suggest that the RF algorithm intrinsically contains a part of exploration due to the bagging paradigm. It could be interesting to add experiments in the future to test this point. 

Still with regard to the random forests, an open question is to study if a better strategy than margin sampling could be designed. Since the random forests are ensemble classifiers, a possible way of research to design this strategy is to check if they could be used in the credal uncertainty framework~\cite{Antonucci2012ActiveLB} which seeks to differentiate between the reducible and irreducible part of the uncertainty in a prediction.

\paragraph{About error generalization.}In Real world application AL should be used most of the time in absence of a test dataset. A open question could be to a use another known result about RF: the possibility to have an estimate of the generalization error using the Out-Of-Bag (OOB) samples \cite{hastie2009,oob}. We did not present experiments on this topic in this paper but an idea could be to analyze the convergence versus the number of labelled examples between the OOB performance and the test performance to check at which ``moment'' ($|L|$) one could trust\footnote{Since when $|L|$ is very low the RF do overtraining thus it's train performance is not a good indicator for the error generalization} the OOB performance  (OOB performance $\approx$ test performance). The use of a ``random uniform forest'' \cite{ciss2015} for which the OOB performance seems to be more reliable could also be investigated.



\paragraph{About the benchmarking methodology.}
Recent benchmarks have highlighted the need for extensive experimentation to compare active learning strategies. The research community might benefit from a ``reference'' benchmark, as in the field of time series classification~\cite{UCRArchive}, so that new results can be rigorously compared to the state of the art on a same and large set of datasets. 
By this way, one will have comprehensive benchmarks that could ascertain the transferability of the learned strategies and demonstrate that these strategies can safely be used in real-world settings.

If this reference benchmark is created, the second step would be to decide how to compare the AL strategies. This comparison could be made using not a single criterion but a ``pool" of criteria. This pool may be chosen to reflect different ``aspects" of the results~\cite{Kottke2017}.





\bibliographystyle{splncs04}
\bibliography{ref}

\begin{thebibliography}{10}
\providecommand{\url}[1]{\texttt{#1}}
\providecommand{\urlprefix}{URL }
\providecommand{\doi}[1]{https://doi.org/#1}

\bibitem{Aggarwal2014}
Aggarwal, C.C., Kong, X., Gu, Q., Han, J., Yu, P.S.: {Active Learning: A
  Survey}. In: Aggarwal, C.C. (ed.) {Data Classification: Algorithms and
  Applications}, chap.~22, pp. 571--605. CRC Press (2014)

\bibitem{Antonucci2012ActiveLB}
Antonucci, A., Corani, G., Bernaschina, S.: {Active Learning by the Naive
  Credal Classifier}. In: Proceedings of the Sixth European Workshop on
  Probabilistic Graphical Models (PGM). pp. 3--10 (2012)

\bibitem{Baram2004}
Baram, Y., El-Yaniv, R., Luz, K.: {Online Choice of Active Learning
  Algorithms}. Journal of Machine Learning Research  \textbf{5},  255--291
  (2004)

\bibitem{Beyer2015}
Beyer, C., Krempl, G., Lemaire, V.: {How to Select Information That Matters: A
  Comparative Study on Active Learning Strategies for Classification}. In:
  Proceedings of the 15th International Conference on Knowledge Technologies
  and Data-driven Business. ACM (2015)

\bibitem{oob}
Breiman, L.: Out-of-bag estimation (1996),
  \url{https://www.stat.berkeley.edu/~breiman/OOBestimation.pdf}, last visited
  08/03/2020

\bibitem{Candillier2013}
{Candillier}, L., {Lemaire}, V.: {Design and analysis of the nomao challenge
  active learning in the real-world.} In: Proceedings of the ALRA: Active
  Learning in Real-world Applications, Workshop ECML-PKDD. (2012)

\bibitem{UCRArchive}
Chen, Y., Keogh, E., Hu, B., Begum, N., Bagnall, A., Mueen, A., Batista, G.:
  {The UCR Time Series Classification Archive} (2015),
  \url{www.cs.ucr.edu/~eamonn/time_series_data/}

\bibitem{Chu2016}
Chu, H.M., Lin, H.T.: {Can Active Learning Experience Be Transferred?} 2016
  IEEE 16th International Conference on Data Mining pp. 841--846 (2016)

\bibitem{ciss2015}
Ciss, S.: {Generalization Error and Out-of-bag Bounds in Random (Uniform)
  Forests}, working paper or preprint,
  \url{https://hal.archives-ouvertes.fr/hal-01110524/document}, last visited
  06/03/2020

\bibitem{Collet2018}
Collet, T.: {Optimistic Methods in Active Learning for Classification}. Ph.D.
  thesis, {Universit{\'e} de Lorraine} (2018)

\bibitem{Dua:2019}
Dua, D., Graff, C.: {UCI Machine Learning Repository} (2017),
  \url{http://archive.ics.uci.edu/ml}

\bibitem{Ebert2012}
Ebert, S., Fritz, M., Schiele, B.: {Ralf: A reinforced active learning
  formulation for object class recognition}. In: 2012 IEEE Conference on
  Computer Vision and Pattern Recognition. pp. 3626--3633 (2012)

\bibitem{Ertekin2007}
Ertekin, S., Huang, J., Bottou, L., Giles, L.: {Learning on the Border: Active
  Learning in Imbalanced Data Classification}. In: Conference on Information
  and Knowledge Management. pp. 127--136. CIKM (2007)

\bibitem{Guyon2011}
Guyon, I., Cawley, G., Dror, G., Lemaire, V.: {Results of the Active Learning
  Challenge}. In: Proceedings of Machine Learning Research. vol.~16, pp.
  19--45. PMLR (2011)

\bibitem{Hasselt2016}
Hasselt, H.v., Guez, A., Silver, D.: {Deep Reinforcement Learning with Double
  Q-Learning}. In: AAAI Conference on Artificial Intelligence. pp. 2094--2100
  (2016)

\bibitem{hastie2009}
Hastie, T., Tibshirani, R., Friedman, J.: The elements of statistical learning:
  data mining, inference and prediction. Springer, 2 edn. (2009)

\bibitem{Hsu2015}
Hsu, W.N., Lin, H.T.: {Active Learning by Learning}. In: Proceedings of the
  Twenty-Ninth AAAI Conference on Artificial Intelligence. pp. 2659--2665. AAAI
  Press (2015)

\bibitem{hllermeier2019aleatoric}
Hüllermeier, E., Waegeman, W.: {Aleatoric and Epistemic Uncertainty in Machine
  Learning: An Introduction to Concepts and Methods}. arXiv\string:1910.09457
  [cs.LG]  (2019)

\bibitem{Konyushkova2017}
Konyushkova, K., Sznitman, R., Fua, P.: {Learning Active Learning from Data}.
  In: Advances in Neural Information Processing Systems 30, pp. 4225--4235
  (2017)

\bibitem{Konyushkova2018}
Konyushkova, K., Sznitman, R., Fua, P.: {Discovering General-Purpose Active
  Learning Strategies}. arXiv\string:1810.04114 [cs.LG]  (2019)

\bibitem{Korner2006}
K\"{o}rner, C., Wrobel, S.: {Multi-class Ensemble-Based Active Learning}. In:
  Proceedings of the 17th European Conference on Machine Learning. pp.
  687--694. Springer-Verlag (2006)

\bibitem{Kottke2017}
Kottke, D., Calma, A., Huseljic, D., Krempl, G., Sick, B.: {Challenges of
  Reliable, Realistic and Comparable Active Learning Evaluation}. In:
  Proceedings of the Workshop and Tutorial on Interactive Adaptive Learning.
  pp. 2--14 (2017)

\bibitem{Lemke2015}
Lemke, C., Budka, M., Gabrys, B.: {Metalearning: a survey of trends and
  technologies}. Artificial Intelligence Review  \textbf{44},  117--130 (2015)

\bibitem{Mnih2013}
Mnih, V., Kavukcuoglu, K., Silver, D., Graves, A., Antonoglou, I., Wierstra,
  D., Riedmiller, M.: {Playing Atari with Deep Reinforcement Learning}.
  arXiv\string:1312.5602 [cs.LG]  (2013)

\bibitem{Nguyen2019EpistemicUS}
Nguyen, V.L., Destercke, S., H{\"u}llermeier, E.: {Epistemic Uncertainty
  Sampling}. In: Discovery Science (2019)

\bibitem{Pang2018}
Pang, K., Dong, M., Wu, Y., Hospedales, T.M.: {Dynamic Ensemble Active
  Learning: A Non-Stationary Bandit with Expert Advice}. In: Proceedings of the
  24th International Conference on Pattern Recognition. pp. 2269--2276 (2018)

\bibitem{Pang2018b}
Pang, K., Dong, M., Wu, Y., Hospedales, T.M.: {Meta-Learning Transferable
  Active Learning Policies by Deep Reinforcement Learning}.
  arXiv\string:1806.04798 [cs.LG]  (2018)

\bibitem{Santos2014}
Pereira-Santos, D., de~Carvalho, A.C.: {Comparison of Active Learning
  Strategies and Proposal of a Multiclass Hypothesis Space Search}. In:
  Proceedings of the 9th International Conference on Hybrid Artificial
  Intelligence Systems -- Volume 8480. pp. 618--629. Springer-Verlag (2014)

\bibitem{PereiraSantos2019}
Pereira-Santos, D., Prudêncio, R.B.C., de~Carvalho, A.C.: {Empirical
  investigation of active learning strategies}. Neurocomputing
  \textbf{326--327},  15--27 (2019)

\bibitem{Pupo2018StatisticalCO}
Pupo, O.G.R., Altalhi, A.H., Ventura, S.: Statistical comparisons of active
  learning strategies over multiple datasets. Knowl. Based Syst.  \textbf{145},
   274--288 (2018)

\bibitem{RamirezLoaiza2017}
Ramirez-Loaiza, M.E., Sharma, M., Kumar, G., Bilgic, M.: {Active learning: an
  empirical study of common baselines}. Data Mining and Knowledge Discovery
  \textbf{31}(2),  287--313 (2017)

\bibitem{Salperwyck2011}
Salperwyck, C., Lemaire, V.: {Learning with few examples: an empirical study on
  leading classifiers}. In: Proceedings of the 2011 International Joint
  Conference on Neural Networks. pp. 1010--1019. IEEE (2011)

\bibitem{Schaul2016}
Schaul, T., Quan, J., Antonoglou, I., Silver, D.: {Prioritized Experience
  Replay}. arXiv\string:1511.05952 [cs.LG]  (2016)

\bibitem{Scheffer2001}
Scheffer, T., Decomain, C., Wrobel, S.: {Active Hidden Markov Models for
  Information Extraction}. In: Hoffmann, F., Hand, D.J., Adams, N., Fisher, D.,
  Guimaraes, G. (eds.) Advances in Intelligent Data Analysis. pp. 309--318
  (2001)

\bibitem{Schein2007}
Schein, A.I., Ungar, L.H.: {Active learning for logistic regression: an
  evaluation}. Machine Learning  \textbf{68},  235--265 (2007)

\bibitem{Settles2012}
Settles, B.: {Active Learning}. Morgan \& Claypool Publishers (2012)

\bibitem{Settles2008}
Settles, B., Craven, M.: {An Analysis of Active Learning Strategies for
  Sequence Labeling Tasks}. In: Proceedings of the Conference on Empirical
  Methods in Natural Language Processing. pp. 1070--1079. Association for
  Computational Linguistics (2008)

\bibitem{Shao2019}
Shao, J., Wang, Q., Liu, F.: {Learning to Sample: An Active Learning
  Framework}. IEEE International Conference on Data Mining (ICDM) pp. 538--547
  (2019)

\bibitem{Trittenbach2018}
Trittenbach, H., Englhardt, A., B{\"{o}}hm, K.: An overview and a benchmark of
  active learning for one-class classification. CoRR  \textbf{abs/1808.04759}
  (2018)

\bibitem{Watkins1992}
Watkins, C.J.C.H., Dayan, P.: {Q-learning}. Machine Learning  \textbf{8}(3),
  279--292 (1992)

\bibitem{Yang2018}
Yang, Y., Loog, M.: {A benchmark and comparison of active learning for logistic
  regression}. Pattern Recognition  \textbf{83},  401--415 (2018)

\end{thebibliography}

\end{document}